\documentclass{article}

\usepackage{arxiv}

\usepackage[utf8]{inputenc} 
\usepackage[T1]{fontenc}    
\usepackage{hyperref}       
\usepackage{url}            
\usepackage{booktabs}       
\usepackage{amsfonts}       
\usepackage{nicefrac}       
\usepackage{microtype}      
\usepackage{lipsum}		
\usepackage{graphicx}
\usepackage{natbib}
\usepackage{doi}
\usepackage{amsmath}

\title{A Fusion Approach of Dependency Syntax and Sentiment Polarity for Feature Label Extraction in Commodity Reviews}

\date{} 					

\author{ \href{https://orcid.org/0009-0006-3794-2802}{\includegraphics[scale=0.06]{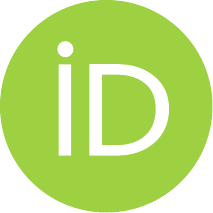}\hspace{1mm}Jianfei~Xu}\thanks{Funding Project: National Undergraduate Innovation and Entrepreneurship Training Program. Project Title: Research on Online Review Analysis System Based on Big Data. Project Number: 201810856013.} \\
	Department of Management\\
	Shanghai University of Engineering Science\\
	Shanghai, China 201620 \\
	\texttt{mr.xujianfei@gmail.com} \\
}

\renewcommand{\headeright}{}
\renewcommand{\undertitle}{}
\renewcommand{\shorttitle}{}

\hypersetup{
pdftitle={A Fusion Approach of Dependency Syntax and Sentiment Polarity for Feature Label Extraction in Commodity Reviews},
pdfsubject={cs.AI},
pdfauthor={Jianfei Xu},
pdfkeywords={Dependency Parsing, Sentiment Analysis, Feature Tagging},
}

\begin{document}
\maketitle

\begin{abstract}

This study uses 13,218 review data entries for four categories of products—mobile phones, computers, cosmetics, and food—from JD.com as the data source. By integrating dependency parsing and sentiment polarity analysis, a novel method for extracting feature tags from product reviews is proposed. This approach not only addresses the issue of low robustness in extraction algorithms but also improves the accuracy of the extraction results. Experimental results show that the accuracy of the proposed extraction method stabilizes at 0.7, with recall and F-score both stabilizing at 0.8, indicating overall satisfactory outcomes. However, issues such as dependence on the matching dictionary and the narrow scope of tag extraction require further investigation.

\end{abstract}


\section{Introduction}

In recent years, e-commerce has experienced rapid development alongside the evolution of internet technologies. On May 31, 2018, the E-commerce Department of the Ministry of Commerce of China released the 2017 China E-commerce Report, highlighting that in 2017, the rapid development of "Internet Plus" accelerated the integration of online and offline channels. E-commerce has become a key driver of China’s economic transformation and upgrading, bringing significant commercial value. According to the report, China’s e-commerce transaction volume in 2017 reached 29.16 trillion yuan, an increase of 11.7\% year-on-year, and 2.8 times the transaction volume of 2013. Notably, online retail sales reached 7.18 trillion yuan, growing 39.1\% compared to 2016. Behind this massive sales figure, e-commerce platforms retain a large amount of user review data. How to extract valuable insights from these review datasets and utilize such information to provide more precise services to users has become a key focus for both industry and academia. \cite{zhao2018} 

Due to the massive volume and high randomness of user review data, information overload often occurs, limiting the effectiveness of e-commerce user reviews as online word-of-mouth. For users, processing a large number of reviews makes it difficult to clearly identify product features, thereby hindering accurate decision-making. For e-commerce platforms, overloaded information obstructs the identification of competitive advantages, the implementation of customer segmentation, and efforts to encourage user participation in online reviews, ultimately reducing the marketing efficiency of word-of-mouth dissemination. Therefore, tools are urgently needed to assist consumers in processing extensive review data by extracting valuable content from the flood of product reviews. These tools should condense the main points and perspectives of reviews into concise, structured feature descriptions that are convenient for both consumers and businesses to use.

To address these issues, feature tagging based on product reviews has emerged as a solution. It extracts reviews in the form of "product feature-evaluation term" pairs. Scholars, drawing on schema theory and cognitive adaptation theory, have studied the effectiveness of feature tags. Their findings suggest that feature tags enhance users' perceived usefulness and satisfaction while mitigating the problem of information overload in reviews \cite{liu2016}. Thus, this study focuses on the effectiveness of feature tags, utilizing dependency parsing theory to extract the correspondence between "product features" and "evaluation terms." By integrating review content, it aims to refine users' opinions on product details and generate feature-based product descriptions.

\section{Related Work}

In the context of big data, feature tags can aggregate user opinions, improve the quality of review content, and help users efficiently extract highly informative and valuable references from online product reviews. This functionality has attracted significant attention from scholars.

To address the issue of correlations among multiple feature tags, scholars have primarily used methods such as information entropy and mutual information. \cite{li2017} improved the traditional k-nearest neighbor (k-NN) multi-label learning algorithm, which often ignores correlations between tags. By using mutual information to measure the knowledge content of tag classification and assigning corresponding weight coefficients, they incorporated tag correlations into the weight coefficients of features, achieving promising experimental results. \cite{chai2018} applied granulation to the tag space based on the optimal granulation number obtained through average information entropy. They used membership degrees to assess the strength of correlations among tags and weighted the features accordingly, addressing issues of feature-to-tag correlation and the combinatorial explosion of tags.

In the issue of multi-feature tag selection, \cite{chen2018} proposed a sparse regularization method based on correlation entropy and feature manifold learning to address the problem of multi-label feature selection through correlations. \cite{zhang2018a} proposed a forward-search-based nonlinear feature selection algorithm, utilizing the theories of mutual information and interaction information to identify the optimal subset related to multi-class labels and reduce computational complexity. \cite{yan2019} established a multi-label feature selection model using a global linear regression function, combined with an h-graph to obtain local descriptive information and improve model accuracy. They introduced \(L_{1,2}\) constraints to enhance the distinguishability between features and the stability of regression analysis, effectively avoiding noise interference.

In the field of feature tag classification, \cite{zhang2019} proposed a multi-label class attribute feature extraction algorithm, LIFT\_RSM, to address the issue of redundancy in the attribute space caused by the construction of class attributes, thereby improving classification performance. \cite{li2014} introduced a multi-label classification algorithm based on information gain. This algorithm initially assumes independence among features and, after calculating mutual information, eliminates irrelevant features based on a threshold to obtain the optimal feature subset.

In the field of feature tag extraction, \cite{bao2018} proposed a review mining model based on domain ontology to identify "feature-opinion pairs" in fresh product reviews, addressing the issue of the rapidly increasing volume of review data. This approach significantly improved computational performance. However, the model's applicability was limited due to the relatively narrow range of data sources. \cite{yin2019} introduced a feature-opinion pair extraction method based on a semantic dictionary, where the matching and extraction algorithm automatically generates feature-opinion pairs.

Drawing on the aforementioned research, this study combines dependency syntax analysis and sentiment polarity methods to propose a method for extracting feature tags from product review corpora. This approach can structurally extract review objects, degree adverbs, negation words, and opinion words, thereby improving the precision and robustness of sentiment word and evaluation object identification. Additionally, through the use of window values, the method enables context analysis of reviews based on sentiment polarity, effectively addressing the ambiguity caused by the one-sided judgments of conventional methods.

\section{Feature Extraction of Product Reviews Based on the Integration of Dependency Syntax and Sentiment Polarity}

Based on the theoretical framework of dependency syntax and the structure of feature tags, each dependency relationship should include at least two components: the core word and the dependent word. Together with the semantic relationship between them, these elements form an evaluative opinion. Building on the traditional "product feature-evaluation term" structure, this study incorporates elements of sentiment analysis. By determining sentiment polarity, negation words and adverbs that indicate the degree of sentiment are integrated into the traditional structure, resulting in a dependency structure of "product feature-negation word-adverb-evaluation term." Accordingly, the process of extracting evaluation features in this study is divided into two parts: first, constructing extraction rules for product features (evaluation objects) and evaluation terms; and second, extracting negation words and adverbs that characterize the degree of sentiment between product features and evaluation terms.

\subsection{Feature Tag Structure}

The feature tags of product reviews incorporating sentiment polarity consist of three components: product feature, sentiment degree, and evaluation term, as shown in Table \ref{table:structure_of_feature_tags}. Among them, the product feature refers to the evaluation subject (i.e., evaluation object) extracted from the review, such as battery life, screen, charging time, appearance, etc. Sentiment terms indicate the sentiment polarity of the product feature as expressed by the reviewer and can be determined jointly by degree adverbs and negation words in the review. For example, in the sentence "The shape of the mouse is very novel, resembling an Apple machine, and its appearance feels very cool," the adverb "very" indicates the degree of sentiment. The evaluation term defines the attitude toward the subject of the product. These three components together form the product feature tag, with sentiment terms and evaluation terms often needing to appear together to reflect the reviewer’s opinion and emotional inclination toward a particular product feature.

\begin{table}[htbp]
    \centering
    \begin{tabular}{llll}
        \toprule
        \textbf{Product Feature} & \textbf{Negation Word} & \textbf{Adverb} & \textbf{Evaluation Word} \\
        \midrule
        Screen & — & Very & Beautiful \\
        Price & Not & — & High \\
        \bottomrule
    \end{tabular}
    \caption{Structure of Feature Tags}
    \label{table:structure_of_feature_tags}
\end{table}

\subsection{Feature Tag Extraction Rules}

Based on the research and analysis of dependency rules for product features and evaluation terms by \cite{nie2014}, this study extends the scope of dependency relationships in reviews. By examining the semantic modification relationships between words, the dependency relationships among "product feature-sentiment term-evaluation term" can be identified and extracted. The main dependency rules are defined as follows:

\begin{itemize}
    \item Search for components dependent on the evaluation term to its left. If the component’s part of speech is an adverb and it functions as an adverbial modifier of the evaluation term, then the sentiment term is the adverbial component modifying the evaluation. For example, in the sentence "The progress is very fast," the word "very" not only modifies "fast" but also serves as an adverbial component of "fast" in the sentence. Therefore, "very" is the adverb modifying the evaluation term "fast."
    \item When both the product feature term and the evaluation term are not core words in the sentence, and the intermediate grammatical element reverses the opinion rather than negating an adverb in the sentence, that word is considered a negation term. For example, in the sentence "The phone is not very good-looking," both "phone" and "good-looking" are not core words, and the intermediate grammatical element "not" negates the opinion about whether the phone is good-looking. Thus, "not" is the negation term for this feature.
\end{itemize}

\subsection{Determination and Filtering of Sentiment Tag Polarity}

Due to issues such as redundancy of sentiment words and low polarity of some sentiment words in the extracted results, this study adopts the word sentiment polarity calculation method proposed by \cite{wang2012}, which integrates HowNet and PMI. This method calculates the polarity of sentiment tags and filters out sentiment words without significant sentiment polarity by setting a threshold.

\begin{itemize}
    \item Assume that the word to compute the polarity of is word\underline{i}. First, use HowNet to perform synonym expansion. Let the expanded synonym set be \( \{ word_1, word_2, word_3, \dots, word_r \} \).
    \item The sentiment polarity of $word_i$ and its synonyms can be calculated using the formula:
    \[
    \text{hownet}(word_i) = \sum_{i=1}^n \text{Sim}(word_i, \text{commendatory}_i) - \sum_{i=1}^n \text{Sim}(word_i, \text{derogatory}_i),
    \]
    where $\text{Sim}(word_i, \text{commendatory}_i)$ represents the similarity between $word_i$ and the commendatory word $\text{commendatory}_i$, and $\text{Sim}(word_i, \text{derogatory}_i)$ represents the similarity between $word_i$ and the derogatory word $\text{derogatory}_i$.
    When the similarity $\text{Sim}(word_i, \text{reference})$ between $word_i$ and the reference word is sufficiently high, the sentiment word $word_i$ can be replaced by the reference word. The similarity $\text{Sim}(word_i, \text{reference})$ is compared with the maximum similarity $Max$, and a predefined threshold $\theta_{\text{HowNet}}$. If $Max > \theta_{\text{HowNet}}$, $word_i$ is considered to have the same sentiment orientation as the reference word.
    The sentiment polarity of $word_i$ is calculated using the formula:
    \[
    \text{SoHowNet}(word_i) = \text{Sim}(word_i, \text{reference}) \times \text{So}(\text{reference}),
    \]
    where $\text{So}(\text{reference})$ denotes the predefined sentiment intensity of the reference word.
    \item $PIM(word_i)$ is calculated as $PIM(word_i) = \sum_{i=1}^n PIM(word_i, \text{commendatory}_i) - \sum_{i=1}^n PIM(word_i, \text{derogatory}_i)$, where $PIM(word_1, word_2) = \frac{P(word_1) \land P(word_2)}{P(word_1)P(word_2)}$, and $P(word_i)$ represents the frequency of $word_i$ appearing independently in the corpus. $PIM(word_i)$ is compared with a predefined threshold $\theta_{PIM}$. If $PIM(word_i) > \theta_{PIM}$, it is considered commendatory; otherwise, it is considered derogatory. The identified words are then added to the set $\{word_1, word_2, word_3, \dots, word_r\}$, and the $PIM$ value of the set is calculated.
    \item Based on the study by \cite{wang2012}, the thresholds $\theta_{\text{HowNet}}$ and $\theta_{\text{PMI}}$ are set to 0.73 and 0.50, respectively. Sentiment words that are below both thresholds are filtered out.
\end{itemize}

\subsection{Feature Tag Extraction}

\begin{figure}[htbp]
    \centering
    \includegraphics[width=1\linewidth]{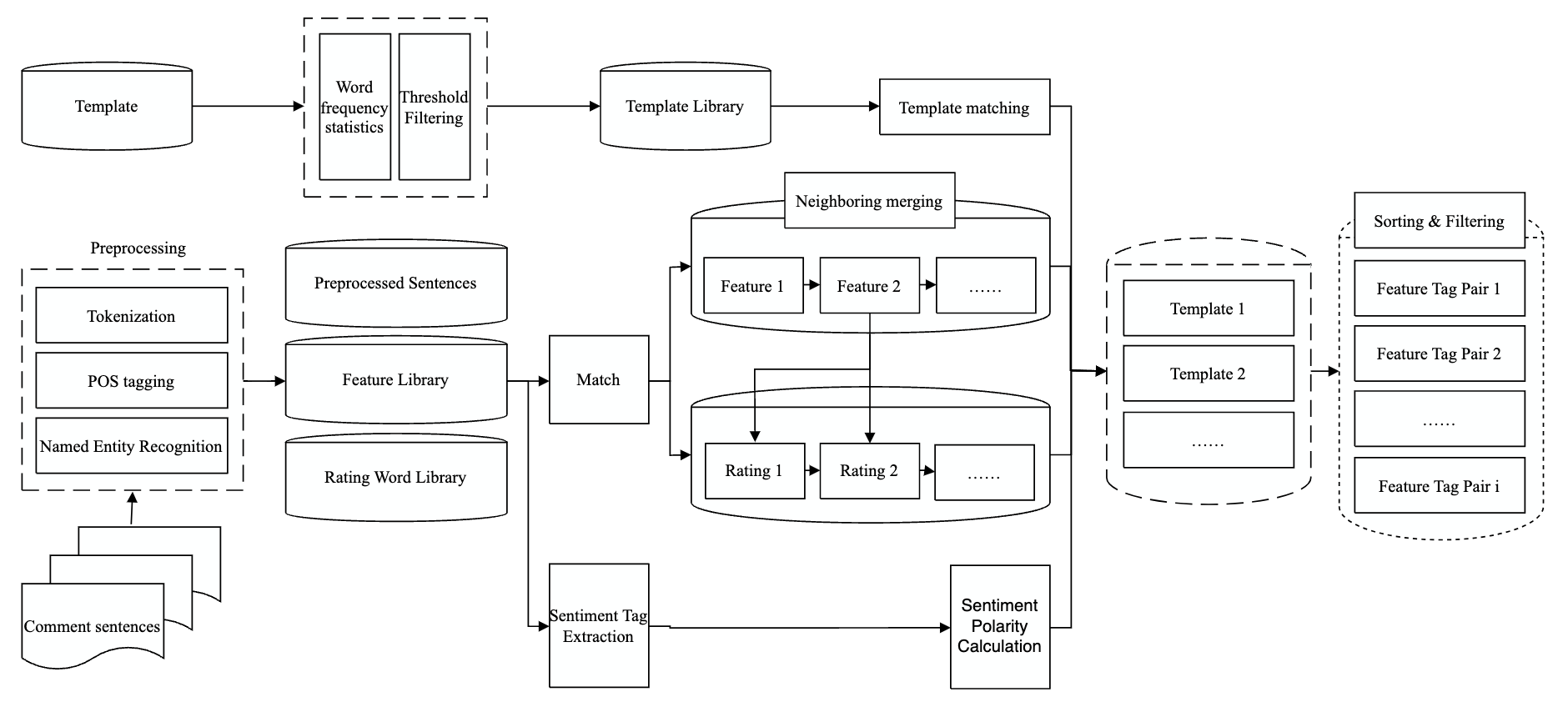}
    \caption{Feature Label Extraction Process}
    \label{fig:feature_label_extraction_process}
\end{figure}

Figure \ref{fig:feature_label_extraction_process} shows the feature tag extraction process in this study, with the specific steps as follows:

\begin{enumerate}
    \item Preprocess the review text, including tokenization, part-of-speech tagging, named entity recognition, and syntactic analysis.
    \item Match the preprocessed sentences with the feature library and rating term library to identify product features, sentiment terms, and evaluation terms.
    \item Since there may be one-to-many or many-to-many relationships between the extracted product features and evaluation terms, it is necessary to combine and merge adjacent product feature terms and evaluation terms. Then, simplify the extracted combinations based on direct and indirect dependency relationships: for directly dependent feature combinations, no restrictive conditions are applied; for indirectly associated combinations, a window value constraint is set, where the specific window value is determined by the length of the syntactic template.
    \item Compare the simplified combinations with the template library, verify them from the perspectives of structural dependency relationships and part of speech, and list combinations that match the template pattern as candidate feature tag pairs.
    \item Rank the filtered candidate feature tag pairs by frequency, and select the feature tags generated by the highest-frequency template as the final extraction results.
\end{enumerate}

\section{Empirical Analysis}

\subsection{Data Source}

The data used in this experiment comes from 13,218 review comments on four types of products—mobile phones, computers, beauty products, and food—from the JD.com online marketplace. The "Jieba" Python toolkit was used for tokenization and part-of-speech tagging of all the review data. Named entity recognition was performed using the Tencent Cloud AI platform. A subset of the review data was manually annotated for product feature terms and evaluation terms. Additionally, 245 sentiment words were referenced from the Baidu Baike encyclopedia for sentiment analysis.

\subsection{Evaluation Metrics}

This study uses accuracy, recall, and F-value to quantitatively evaluate the performance of the experimental results. The definitions of each metric are as follows:

\begin{itemize}
    \item Accuracy: Represents the ratio of correctly extracted feature tags to the total number of extracted feature tags. 
    \[
    P = \frac{A}{A + B} \tag{1}
    \]
    Where \(P\) is accuracy, \(A\) is the number of correctly extracted feature tags, and \(B\) is the number of incorrectly extracted feature tags.
    \item Recall: Represents the ratio of correctly extracted sentiment tags to the total number of feature tags that actually exist in the opinion corpus. 
    \[
    R = \frac{A}{A + C} \tag{2}
    \]
    Where \(R\) is recall, \(A\) is the number of correctly extracted sentiment tags, and \(C\) is the number of feature tags that are missing in the extraction process. Where \( R \) is the recall, \( A \) is the number of correctly extracted feature tags, and \( C \) is the number of missed feature tags.
    \item F-value: It is used to comprehensively measure both accuracy and recall metrics.
    \[
    F = \frac{2PR}{P + R} \tag{3}
    \]
\end{itemize}

\subsection{Results Analysis}

\subsubsection{Comparison and Analysis of Feature Tag Extraction Results}

By analyzing the results in Table \ref{table:feature_extraction_results_based_on_dependency_syntax} and Table 3, it can be observed that this study is capable of fully capturing the sentiment orientation in user reviews. For example, in review number 2124, the original meaning of the sentence is that the taste is not crispy. However, because the extraction method based on dependency syntax does not consider semantic issues and relies solely on syntactic structure, the result extracted was "crispy taste," leading to an incorrect feature tag extraction.

\begin{table}[htbp]
    \centering
    \begin{tabular}{lp{15em}lll} 
        \toprule
        No. & Review Text & Target Object & Negation Word & Adverb Evaluation Word \\ 
        \midrule
        2132  & The price is very affordable, limited-time offer & Price & - & Affordable \\
        352   & The appearance is almost indistinguishable from the fake, please no complaints from authentic product users & Appearance & - & Almost indistinguishable \\
        2124  & The crispy texture is not as crisp as expected, disappointing shopping experience & Taste & Not & Crispy \\
        11850 & The D of the computer uses skin-like material, feels great, great value for money & Feel & - & Great \\
        \bottomrule
    \end{tabular}
    \caption{Feature Extraction Results Based on Dependency Syntax}
    \label{table:feature_extraction_results_based_on_dependency_syntax}
\end{table}

\begin{table}[htbp]
    \centering
    \begin{tabular}{lp{13em}llll}
        \toprule
        No. & Review Text & Target Object & Negation Word & Adverb & Evaluation Word \\ 
        \midrule
        2132 & The price is very affordable, limited-time offer & Price & None & - & Affordable \\
        352  & The appearance is almost indistinguishable from fake, no complaints from authentic product users & Appearance & - & Enough & Almost indistinguishable \\
        2124 & The crispy texture is not as crispy as expected, disappointing shopping experience & Taste & Not & - & Crispy \\
        11850 & The computer's D uses skin-like material, feels great, great value for money & Feel & - & Very & Great \\
        \bottomrule
    \end{tabular}
\caption{Feature Label Extraction Results}
\label{table:feature_label_extraction_results}
\end{table}

\subsubsection{Analysis of Evaluation Metrics}

From Tables \ref{table:feature_extraction_evaluation_metrics_based_on_dependency_syntax} and \ref{table:evaluation_metrics_analysis_results}, it can be seen that, using the text-based method, the accuracy of feature extraction for mobile phones, computers, beauty products, and food categories is generally around 0.7, with recall rates around 0.8. Overall, the results are better than those of feature extraction based on dependency syntax. In terms of accuracy, the food category has the highest accuracy. This is because user reviews of food products are generally less technical than those for mobile phones and computers, and less verbose than beauty product reviews. The data is relatively clean and well-organized, leading to simpler sentence structures and more precise word expressions. In terms of recall, all four categories show high recall rates, indicating that the feature extraction method in this study is capable of identifying certain grammatical structures in the reviews. Regarding the F-value, it remains stable at around 0.8, suggesting that the overall extraction performance is relatively ideal and meets the expectations of the experiment.

\begin{table}[htbp]
    \centering
    \begin{tabular}{lcccccccc} 
        \toprule
        Product Category & Total Extracted & Correct Extracted & Unidentified & Accuracy (\%) & Recall (\%) & F-Score \\
        \midrule
        Phone    & 4322 & 2426 & 301 & 0.56 & 0.89 & 0.69 \\
        Computer & 5217 & 3432 & 120 & 0.66 & 0.97 & 0.79 \\
        Beauty   & 1244 & 563  & 461 & 0.45 & 0.55 & 0.50 \\
        Food     & 7001 & 5329 & 598 & 0.76 & 0.90 & 0.82 \\
        \midrule
        Total    & 17784 & 12247 & 1463 & 0.69 & 0.89 & 0.78 \\
        \bottomrule
    \end{tabular}
    \caption{Feature Extraction Evaluation Metrics Based on Dependency Syntax}
    \label{table:feature_extraction_evaluation_metrics_based_on_dependency_syntax}
\end{table}

\begin{table}[htbp]
    \centering
    \begin{tabular}{lcccccccc} 
        \toprule
        Product Category & Total Extracted & Correct Extracted & Unidentified & Accuracy (\%) & Recall (\%) & F-Score \\
        \midrule
        Phone    & 4322 & 3124 & 342 & 0.72 & 0.90 & 0.80 \\
        Computer & 5217 & 4023 & 122 & 0.77 & 0.97 & 0.86 \\
        Beauty   & 1244 & 792  & 462 & 0.64 & 0.63 & 0.63 \\
        Food     & 7001 & 6321 & 649 & 0.90 & 0.91 & 0.90 \\
        \midrule
        Total    & 17784 & 14260 & 1575 & 0.76 & 0.85 & 0.80 \\
        \bottomrule
    \end{tabular}
    \caption{Evaluation Metrics Analysis Results}
    \label{table:evaluation_metrics_analysis_results}
\end{table}

\section{Conclusion}

This study proposes an effective method for extracting feature tags from product reviews by integrating dependency syntax analysis and sentiment polarity. The proposed method successfully captures both the semantic and syntactic features of product reviews, improving the accuracy and robustness of sentiment analysis. Through empirical analysis, the method demonstrated promising results, with accuracy and recall rates around 0.7 and 0.8, respectively, and a stable F-value of approximately 0.8. Despite these successes, there are still challenges, particularly related to the reliance on syntactic structures that may lead to incorrect feature tag extraction in some cases. Future research could focus on further enhancing the method's ability to handle complex sentence structures and improve the semantic understanding of reviews. Overall, the study provides a solid foundation for improving the quality of feature tag extraction in online product reviews, offering valuable insights for both consumers and e-commerce platforms in enhancing decision-making processes and marketing strategies.

\newpage
\bibliographystyle{unsrtnat}
\bibliography{references} 

\end{document}